\documentclass[10pt]{scrartcl}
\usepackage[protrusion=true,expansion=true]{microtype} 
\usepackage{amsmath,amsfonts,amsthm,amssymb,color,epsfig,fullpage} 
\usepackage{verbatim,fancyvrb}
\usepackage{alltt}
\usepackage{tikz}
\usepackage{enumerate}
\usepackage{caption}
\usepackage{subfigure}
\usepackage[]{algorithm}
\usepackage{algpseudocode}
\usepackage{stmaryrd}
\usepackage{afterpage}
\usepackage{url}
\usepackage{xcolor,colortbl}

\newtheorem{definition}{Definition}[section]

\newtheorem{proposition}[definition]{Proposition}

\def\N{{\mathbb N}}

\def\R{{\mathbb R}}

\DeclareMathOperator*{\argmin}{argmin}
\DeclareMathOperator{\dtw}{\delta}

\newcommand{\commentout}[1]{}

\newcommand{\abs}[1]{\mathop{\left\lvert #1 \right\rvert}} 
\newcommand{\args}[1]{\mathop{\left( #1 \right)}} 

\newcommand{\norm}[1]{\mathop{\left\lVert #1 \right\rVert}}
\newcommand{\cbrace}[1]{\mathop{\left\{ #1 \right\}}}

\newcommand{\argsS}[2]{\mathop{\left( #1 \right)#2}} 

\newcommand{\normS}[2]{\mathop{\left\lVert #1 \right\rVert#2}}

\renewcommand{\S}[1]{{\mathcal{#1}}}           	

\renewenvironment{cases}{%
\left\{\begin{array}{c@{\quad : \quad}l}}%
{%
\end{array}\right.}

\begin{document}

\title{Asymmetric Learning Vector Quantization for Efficient Nearest Neighbor Classification in Dynamic Time Warping Spaces }

\author{Brijnesh J.~Jain and David Schultz\\
       Technische Universit\"at Berlin, Germany\\
       e-mail: brijnesh.jain@gmail.com}
\date{}
\maketitle

\begin{abstract}  
The nearest neighbor method together with the dynamic time warping (DTW) distance is one of the most popular approaches in time series classification. This method suffers from high storage and computation requirements for large training sets. As a solution to both drawbacks, this article extends learning vector quantization (LVQ) from Euclidean spaces to DTW spaces. The proposed LVQ scheme uses asymmetric weighted averaging as update rule. Empirical results exhibited superior performance of asymmetric generalized LVQ (GLVQ) over other state-of-the-art prototype generation methods for nearest neighbor classification. 
\end{abstract}

\section{Introduction}

The nearest neighbor (NN) classifier endowed with the dynamic time warping (DTW) distance is one of the most popular methods in time series classification \cite{Fu2011,Xing2010}. Application examples include electrocardiogram frame classification \cite{Huang2002}, gesture recognition \cite{Alon2009,Reyes2011}, speech recognition \cite{Myers1981}, and voice recognition \cite{Muda2010}.

Two disadvantages of the naive NN method are high storage and computation requirements. Storage requirements are high, because the entire training set needs to be retained for being able to execute its classification rule. Computation requirements are high, because classifying a test example demands calculation of DTW distances between the test and all training examples. One solution to both limitations are data reduction methods that infer a small set of prototypes from the training examples \cite{Wilson2000}. These methods aim at scaling down storage and computational complexity, while maintaining high classification accuracy. Two common reduction approaches are prototype selection \cite{Garcia2012} and prototype generation \cite{Triguero2012}. Prototype selection methods choose a suitable subset of the original training set. Examples of prototype selection algorithms that have been applied in DTW spaces are min-max centroids \cite{Rabiner1978}, k-Medoids \cite{Kaufman1987,Petitjean2014,Petitjean2016} and the DROP-family \cite{Wilson2000,Xi2006}.

Prototype generation methods infer new artificial prototypes from the training examples. We distinguish between three directions to prototype generation methods for NN classification in DTW spaces: 
\begin{enumerate}
\itemsep0em
\item \emph{Unsupervised prototype generation} \cite{Abdulla2003,Ongwattanakul2009,Petitjean2016,Rabiner1978,Rabiner1979,Sathianwiriyakhun2016,Srisai2009,Wilpon1985}:
Unsupervised methods cluster the training examples of every class separately. Centroids of the clusters are computed by averaging warped time series, which is non-trivial as compared to averaging vectors \cite{Petitjean2011,Schultz2017}. The resulting centroids form a reduced set of prototypes for NN classification. 

\item \emph{Symmetric LVQ1} \cite{Somervuo1999}:
Symmetric LVQ1 is a supervised prototype generation method that extends the LVQ1 algorithm \cite{Kohonen2001} from Euclidean to DTW spaces by using a symmetric update rule. Starting with an initial set of prototypes, LVQ1 repeatedly applies the following steps: (i) select the next training example $x$; (ii) identify the prototype $p$ closest to $x$; and (iii) attract $p$ to $x$ if the class labels of both agree, otherwise repel $p$ from $x$. 

\item \emph{Relational LVQ} \cite{Gisbrecht2012,Hammer2011,Hammer2014,Mokbel2015}:
Relational LVQ methods extend state-of-the-art LVQ methods from Euclidean spaces to pseudo-Euclidean spaces via pairwise dissimilarity data. An important example that has been extended to relational learning is generalized LVQ (GLVQ) \cite{Sato1996}
\end{enumerate}

An empirical comparison of different methods across the three directions is missing. Consequently, it is unclear which direction is best suited for which situation. Relational methods are theoretically best developed and the most general approach, because they can be applied to any distance space. Unsupervised methods are currently the most popular direction in DTW spaces. Their usefulness has been first demonstrated in the 1970ies for speech recognition \cite{Rabiner1978,Rabiner1979} and recently confirmed for general time series classification tasks \cite{Petitjean2016,Sathianwiriyakhun2016}. Moreover, empirical result showed that k-means together with the DBA algorithm for time series averaging exhibited the best generalization performance over different prototype selection and unsupervised prototype generation methods \cite{Petitjean2014,Petitjean2016}. A limitation of unsupervised methods is that they learn prototypes of every class separately without considering the decision boundaries to the respective neighboring classes.
In contrast to unsupervised methods, supervised approaches aim at directly approximating the true but unknown decision boundaries. As far as we know, symmetric LVQ1 \cite{Somervuo1999} is the only existing supervised prototype generation method that operates in DTW spaces. However, there are two major issues related to symmetric LVQ1: 
\begin{enumerate}
\itemsep0em
\item
The update rule of the Euclidean LVQ1 method can be formulated as a weighted average of two points. Averages in Euclidean spaces are unique minimizers of a sum of squared Euclidean distances. In DTW spaces there are two forms of averages: a symmetric and an asymmetric form \cite{Kruskal1983,Schultz2017}. The symmetric form computes the arithmetic mean of warped time series, whereas the asymmetric form is a minimizer of a sum of squared DTW distance criterion \cite{Jain2016,Schultz2017}. Symmetric LVQ1 uses a symmetric form of weighted average as update rule, which has the semblance of the Euclidean LVQ1 update rule but resists a theoretical justification. 
\item 
Symmetric LVQ1 has been proposed as a heuristic without explicit link to a cost function. It extends the first and simplest LVQ variant formulated in Euclidean spaces \cite{Kohonen2001}. The Euclidean LVQ1 showed unsatisfactory generalization performance, slow convergence, and numerical instabilities. Improved methods such as GLVQ define explicit cost functions that are minimized by stochastic gradient descent. In principle, it is possible to extend cost-based LVQ methods to DTW spaces using symmetric update rules. However, these update rules will fail to minimize the respective extended cost functions in an analytically justified way (cf.~Section \ref{subsec:symmetric-asymmetric}). 
\end{enumerate}
Given both issues, it seems natural to ask whether asymmetric extensions of cost-based LVQ methods are beneficial for supervised prototype generation in DTW spaces.

In this contribution, we propose an asymmetric LVQ scheme for DTW spaces. The proposed asymmetric scheme is generic and theoretically better grounded than its symmetric counterpart. It is generic in the sense that Euclidean LVQ algorithms can be directly extended to DTW spaces under mild assumptions. As examples, we derive asymmetric versions of LVQ1 and GLVQ. The proposed scheme is theoretically justified in the sense that asymmetric update rules are weighted averages that minimize a sum of squared distance criterion. In addition, asymmetric cost-based LVQ are stochastic gradient descent method almost surely. In experiments, we compared the performance of the proposed asymmetric LVQ methods against prototype generation methods from the above three directions. The results suggest that asymmetric GLVQ best trades classification accuracy against computation time by a large margin.

The implications of this contribution are twofold: First, asymmetric GLVQ is well suited for online settings and in situations where storage and computation requirements are an issue. Second, the generic asymmetric LVQ scheme can serve as a blueprint for directly extending unsupervised prototype learning methods to DTW spaces, such as vector quantization, self-organizing maps, and neural gas. 

The rest of this paper is structured as follows: Section 2 introduces LVQ in Eucldean space. Section 3 proposes asymmetric LVQ for DTW spaces. Section 4 presents and discusses experiments. Finally, Section 5 concludes with a summary of the main results and an outlook to further research.

\section{Learning Vector Quantization}

This section introduces learning vector quantization in Euclidean spaces in such a way that most concepts can be directly extended to other distance spaces.

\subsection{Nearest Neighbor Classification}

Let $\args{\S{X}, \delta}$ be a distance space with distance function $\delta:\S{X} \times \S{X} \rightarrow \R_{\geq 0}$. Furthermore, let $\S{Y} = \cbrace{1, \ldots, C}$ be a set consisting of $C$ class labels. We assume that there is an unknown function 
\[
y : \S{X} \rightarrow \S{Y}, \quad x \mapsto y(x)
\]
that assigns every element $x \in \S{X}$ to a class label $y(x) \in \S{Y}$. A nearest neighbor classifier approximates the unknown function $y$ by using a set $\S{C} = \cbrace{\args{p_1, z_1}, \ldots, \args{p_K, z_K)}}$ of $K$ reference elements $p_k \in \S{X}$ with class labels $z_k = y(p_k)$. The set $\S{C}$ is called \emph{codebook} and its elements $p_k$ are the \emph{prototypes}. We demand that $y(\S{C})  = \S{Y}$, that is there is at least one prototype for every class. For the sake of convenience, we occasionally write $p \in \S{C}$ instead of $(p,z) \in \S{C}$.  Consider the function 
\[
p_*(x) = \argmin_{p \in \S{C}} \delta(p, x)
\]
that associates element $x \in \S{X}$ with its nearest (best-matching) prototype $p_*(x) \in \S{C}$. Then the nearest neighbor classifier with respect to codebook $\S{C}$ is a function $h_{\S{C}}: \S{X} \rightarrow \S{Y}$ of the form $h_{\S{C}}(x) = y\args{p_*(x)}$.
The function $h_{\S{C}}(x)$ assigns element $x$ to the class of its nearest prototype $p_*(x)$.

\subsection{Learning Vector Quantization}
Let $\S{X} = \R^d$ be the $d$-dimensional Euclidean space endowed with the Euclidean metric $\delta(x, y) = \norm{x - y}$. Suppose that
$\S{D} = \cbrace{\args{x_1, y_1}, \ldots, \args{x_N, y_N}} \subseteq \S{X} \times \S{Y}$ is a training set. The goal of LVQ is to learn a codebook $\S{C}$ of size $K \ll N$ on the basis of the training set $\S{D}$ such that the expected misclassification error of the nearest neighbor classifier $h_{\S{C}}$ is as small as possible.

Learning can be performed in batch or incremental (stochastic) mode. Here, we focus on incremental learning. During learning, prototypes $p$ are adjusted in accordance with the distortion
\[
D_x(p) = \delta^2(p, x), 
\] 
where $x$ is the current input example. The distortion $D_x(p)$ is differentiable as a function of $p$ and its gradient is given by 
$\nabla D_x(p) = 2(p-x)$. Then the generic LVQ scheme is of the following form: 

\bigskip

\hrule
\begin{enumerate}
\itemsep0em
\item Initialize codebook $\S{C}$. 
\item Repeat until termination:
\begin{enumerate}
\itemsep0em
\item Randomly select a training example $(x, y) \in \S{D}$.
\item For all prototypes $(p, z) \in \S{C}$ do
\begin{align}\label{eq:generic-Euclidean-update-rule}
p \;\leftarrow\; p - \eta \cdot f(p, x)\cdot(p-x),
\end{align}
where $\eta$ is an adaptive learning rate and $f$ is a class-compatible force function. 
\item Optionally adjust hyper-parameters.
\end{enumerate}
\end{enumerate}
\hrule

\bigskip

The learning rate $\eta$ in Eq.~\eqref{eq:generic-Euclidean-update-rule} absorbs the constant factor $2$ of the gradient $\nabla D_x(p)$. Variants of LVQ algorithms differ in the particular form of the force function. Note that the notation of the force function is incomplete for the sake of simplicity. A force function $f(p,x)$ depends on the class labels of both arguments as well as on the entire codebook $\S{C}$. 

The force function $f(p,x)$ is a real-valued function that determines the direction and relative strength with which prototype $p$ is updated. A positive force $f(p,x)$ moves $p$ closer to the input $x$. A negative force $f(p,x)$ repels $p$ from $x$. Finally, prototype $p$ remains unchanged if the force $f(p,x)$ is zero. We demand that the force function $f(p,x)$ is class-compatible in the sense that $f(p,x) \geq 0$ if $y(x) = y(p)$ and $f(p,x) \leq 0$ if $y(x) \neq y(p)$.

Hyper-parameters need to be initialized and can be optionally adjusted during learning. Two common hyper-parameters are the number $K$ of prototypes and the learning rate $\eta$. Some force functions include additional hyper-parameters.

\subsection{Examples of LVQ Algorithms}
This section presents two examples of LVQ algorithms, the LVQ1 and the generalized LVQ algorithm.

\subsubsection*{Example 1: LVQ1}
The LVQ1 algorithm proposed by Kohonen \cite{Kohonen2001} is historically the first LVQ method. Its heuristically motivated update rule is of the form  
\begin{align*}
p \;\leftarrow\; 
\begin{cases}
p - \eta \args{p - x} & p = p_*(x) \text{ and } y(x) = y(p)\\[0.5ex]
p + \eta \args{p - x}  & p = p_*(x) \text{ and } y(x) \neq y(p)\\[0.5ex]
p  & p \neq p_*(x)
\end{cases}.
\end{align*}
The update rule adjusts the best matching prototype $p = p_*(x)$ of the current input $x$ and leaves all other prototypes unchanged. Prototype $p$ is attracted to $x$ if their class labels agree and repelled otherwise. To express the update rule of LVQ1 in terms of Eq.~\eqref{eq:generic-Euclidean-update-rule}, we define the force function $f$ of LVQ1 as
\[
f(p,x)  = \begin{cases}
+1 & p = p_*(x) \text{ and } y(x) = y(p)\\[0.5ex]
-1 & p = p_*(x) \text{ and } y(x) \neq y(p)\\[0.5ex]
0 &  p \neq p_*(x)
\end{cases}.
\]

\subsubsection*{Example 2: Generalized LVQ}

The early LVQ1 algorithm suffered from sensitivity to initialization, instabilities during learning, and slow convergence \cite{Nova2014}. Therefore, several modifications of LVQ1 have been suggested \cite{Biehl2016,Nova2014}. One example is generalized learning vector quantization (GLVQ) proposed by Sato and Yamada \cite{Sato1996}. During learning, GLVQ maximizes the hypothesis margin \cite{Crammer2002,Hammer2005} by minimizing the cost function
\[
E = \sum_{i=1}^N h(\kappa(x_i)),
\]
where $h: \R \rightarrow \R$ is a monotonously increasing function and $\kappa: \S{X} \rightarrow \R$ is the relative distance difference. 

Here, we assume that $h$ is the sigmoid function $h(u) = 1 / (1 + \exp(-\sigma u))$, where $\sigma$ is an adjustable hyper-parameter that controls the slope. To describe the relative distance difference, we consider a training example $(x, y) \in \S{D}$. Suppose that $p^+$ is the closest prototype of $x$ with $y(p^+) = y$ and $p^-$ is the closest prototype of $x$ with $y(p^-) \neq y$. Then the relative distance difference of $x$ is defined by 
\[
\kappa(x) = \frac{d^+ - d^-}{d^+ + d^-},
\]
where $d^{\pm} = \delta(p^{\pm}, x)$ are the squared distances of $x$ from the prototypes $p^{\pm}$. The relative difference $\kappa(x)$ as a function of the prototypes $p^{\pm}$ has the following analytical properties: 
\begin{enumerate}[(i)]
\itemsep0em
\item $\kappa(x)$ is undefined if $x = p^+ = p^-$.
\item $\kappa(x)$ is locally Lipschitz continuous if $d^+ + d^- \neq 0$.
\item $\kappa(x)$ is non-differentiable if one of the prototypes $p^{\pm}$ is not uniquely determined for $x$.
\end{enumerate}
The function $\kappa(x)$ is not continuously extendable at its discontinuity, which is a singleton. Otherwise, from case (ii) follows that 
$\kappa(x)$ is differentiable almost everywhere by Rademacher's Theorem \cite{Evans1992}. In other words, we have zero probability that non-differentiability occurs. The same analytical properties hold for the loss $h(\kappa(x))$ and the cost function $E$. 

The GLVQ algorithm minimizes the cost function $E$ incrementally according to the following update rule for every prototype $p \in \S{C}$:
\begin{align*}
p \;\leftarrow\; 
\begin{cases}
p - \eta \, \phi^+\args{p - x} & p = p^+ \\[0.5ex]
p + \eta \, \phi^-\args{p - x}  & p = p^- \\[0.5ex]
p  & \text{otherwise}
\end{cases},
\end{align*}
where 
\begin{align*}
\phi^+ = h'(\kappa(x))\frac{d^-}{\argsS{d^+ + d^-}{^2}}
\qquad \text{and} \qquad
\phi^- = h'(\kappa(x))\frac{d^+}{\argsS{d^+ + d^-}{^2}}
\end{align*}
The derivative of the sigmoid function is given by $h'(u) = h(u)\args{1 - h(u)}$. 

The update rule of GLVQ performs stochastic gradient descent if $h(\kappa(x))$ is differentiable as a function of $p$, which is almost surely the case. If the exceptional case (i) occurs, we find that $p-x = 0$. In this case, updating leaves the prototype $p$ unchanged. Thus, the singularity of $\kappa$ has no adverse effects. Critical are continuous but non-differentiable points $p$ corresponding to case (iii). There are two straightforward strategies to cope with non-differentiability: The first strategy ignores the current input $x$ and draws the next training example. The second strategy breaks all ties if the closest prototypes $p^{\pm}$ of input $x$ are not uniquely determined and then proceeds as usual. Either of both strategies can be implemented into the force function of GLVQ. The force function of GLVQ following the first strategy for case (iii) is of the form
\[
f(p,x)  = \begin{cases}
+\phi^+(x) & p = p^+ \text{ and } p^+ \text{ is unique} \\[0.5ex]
-\phi^-(x) & p = p^- \text{ and } p^- \text{ is unique}\\[0.5ex]
0 &  \text{otherwise}
\end{cases}.
\]
Substituting the force function $f(p,x)$ into the generic Euclidean update rule \eqref{eq:generic-Euclidean-update-rule} gives the update rule of GLVQ.

\section{LVQ in DTW Spaces}

This section presents a generic asymmetric update rule for LVQ in DTW spaces and discusses its relationship to the symmetric LVQ1 update rule proposed by Somervuo and Kohonen \cite{Somervuo1999}.

\subsection{The Dynamic Time Warping Distance}\label{subsec:DTW}

\begin{figure}
\centering
\includegraphics[width=0.98\textwidth]{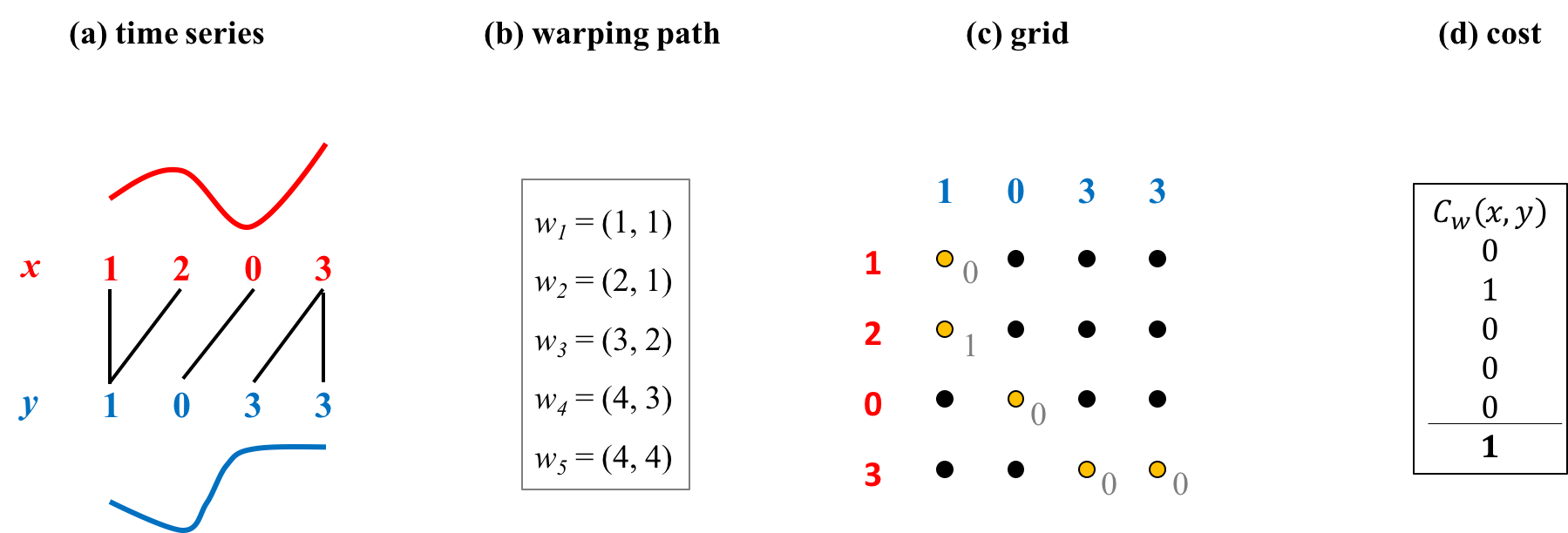}
\caption{\textbf{(a)} Two time series $x$ and $y$ of length $n = 4$. The red and blue numbers are the elements of the respective time series. \textbf{(b)} Warping path $w = (w_1, \ldots, w_5)$ of length $L = 5$. The points $w_l = (i_l, j_l)$ of warping path $w$ align elements $x_{i_l}$ of $x$ to elements $y_{j_l} $ of $y$ as illustrated in (a) by black lines. \textbf{(c)} The $4 \times 4$ grid showing how warping path $w$ moves from the upper left to the lower right corner as indicated by the orange balls. The numbers attached to the orange balls are the squared-error costs of the corresponding aligned elements. \textbf{(d)} The cost $C_w(x, y)$ of aligning $x$ and $y$ along warping path $w$.}
\label{fig:wpath}
\end{figure}

We begin with introducing the DTW distance. We refer to Figure \ref{fig:wpath} for explanatory illustrations of the concepts. 

\medskip

A time series $x$ of length $m$ is a sequence $x = (x_1, \ldots, x_m)$ consisting of feature vectors $x_i \in \R^d$ for every time point $i \in [m] = \cbrace{1, \ldots, m}$. By $\S{T}$ we denote the set of all time series of finite length with features from $\R^d$. To define the DTW-distance, we first need to introduce warping paths.
\begin{definition}
Let $m, n \in \N$. A \emph{warping path} of order $m \times n$ is a sequence $w = (w_1 , \dots, w_L)$ of $L$ points $w_l = (i_l,j_l) \in [m] \times [n]$ such that
\begin{enumerate}
\item $w_1 = (1,1)$ and $w_L = (m,n)$ \hfill\emph{(\emph{boundary conditions})}
\item $w_{l+1} - w_{l} \in \cbrace{(1,0), (0,1), (1,1)}$ for all $l \in [L-1]$ \hfill\emph{(\emph{step condition})} 
\end{enumerate}
\end{definition}
The set of all warping paths of order $m \times n$ is denoted by $\S{W}_{m,n}$. A warping path of order $m \times n$ can be thought of as a path in a $[m] \times [n]$ grid, where rows are ordered top-down and columns are ordered left-right. The boundary conditions demand that the path starts at the upper left corner and ends in the lower right corner of the grid. The step condition demands that a transition from one point to the next point moves a unit in exactly one of the following directions: down, right, and diagonal. 

A warping path $w = (w_1, \ldots, w_L)\in \S{W}_{m,n}$ defines an alignment (or warping) between time series $x = (x_1, \ldots, x_m)$  and $y = (y_1, \ldots, y_n)$. Every point $w_l = (i_l,j_l)$ of warping path $w$ aligns feature vector $x_{i_l}$ to feature vector $y_{j_l}$. Occasionally, we write $\S{W}(x, y)$ instead of $\S{W}_{m,n}$ to denote the set of all warping paths aligning time series $x$ and $y$. 

The \emph{cost} of aligning time series $x$ and $y$ along warping path $w$ is defined by
\begin{equation*}
C_w(x, y) = \sum_{l=1}^L \normS{x_{i_l}-y_{j_l}}{^2},
\end{equation*}
where $\norm{\cdot}$ denotes the Euclidean norm. Then the DTW-distance between two time series minimizes the cost of aligning both time series over all possible warping paths. 
\begin{definition}
The \emph{DTW-distance} between time series $x, y \in \S{T}$ is defined by
\begin{equation*}
\dtw(x, y) = \min \cbrace{\sqrt{C_w(x, y)} \,:\, w \in \S{W}(x,y)}.
\end{equation*}
An \emph{optimal warping path} is any warping path $w \in \S{W}(x, y)$ that satisfies $\dtw(x, y) = \sqrt{C_w(x, y)}$.
\end{definition}
The DTW-distance is not a metric, because it violates the identity of indiscernibles and the triangle inequality.\footnote{The identity of indiscernibles demands that $\delta(x, y) = 0 \Leftrightarrow x = y$ for all $x, y \in \S{T}$.} Instead, the DTW-distance satisfies the following properties for all $x, y \in \S{T}$: (i) $\dtw(x, y) \geq 0$, (ii) $\dtw(x, x) = 0$, and (iii) $d(x, y) = d(y, x)$. Computing the DTW distance and deriving an optimal warping path is usually solved by applying techniques from dynamic programming \cite{Sakoe1978}.

\subsection{A Generic Asymmetric LVQ Update Rule}

This section extends LVQ update rules from Euclidean spaces to DTW spaces using asymmetric averaging.

\medskip

The basic idea for extending LVQ to DTW spaces is as follows: Replace the gradient of the squared Euclidean distance by a suitable surrogate time series $g_{p,x}$ such that the corresponding update rule takes the form
\begin{align}\label{eq:proposed-update-rule}
p' = p - \eta \cdot f(p,x)\cdot g_{p,x},
\end{align}
where $x$ is the current input, $p$ is the prototype to be updated, and $p'$ is the updated prototype. We demand that $g$ has the same length as $p$. In this case, the right hand side of update rule \eqref{eq:proposed-update-rule} can be regarded as a valid algebraic expression of two vectors. 

To construct the surrogate time series $g_{p,x}$, we consider the squared DTW distortion 
\[
D_x: \R^m \rightarrow \R, \quad p \mapsto \dtw^2(p, x),
\]
where $x \in \S{T}$ is a time series of arbitrary length. Restricting the domain of $D_x$ to time series of fixed length $m$ is owed to the principle of asymmetric averaging, which will be discussed later in Section \ref{subsec:symmetric-asymmetric}. Note that different prototypes $p_j$ may have different lengths $m_j$, but the  update $p'_j$ of a prototype $p_j$ has the same length $m_j$ as $p_j$. 

The distortion $D_x$ is a locally Lipschitz continuous function on $\R^m$ \cite{Schultz2017} and therefore differentiable almost everywhere by Rademacher's Theorem \cite{Evans1992}. Every locally Lipschitz continuous function $F: \R^m \rightarrow \R$ admits a concept of generalized gradient $\partial F(p)$ at point $p$, called subdifferential henceforth \cite{Clarke1990}. The subdifferential $\partial F(p) \subseteq \R^m$ of function $F$ is a convex closed subset, whose elements are called subgradients. At differentiable points $p$, the subdifferential $\partial F(p) = \cbrace{\nabla F(p)}$ coincides with the gradient of $F$. 

\begin{figure}[t]
\centering
\includegraphics[width=0.9\textwidth]{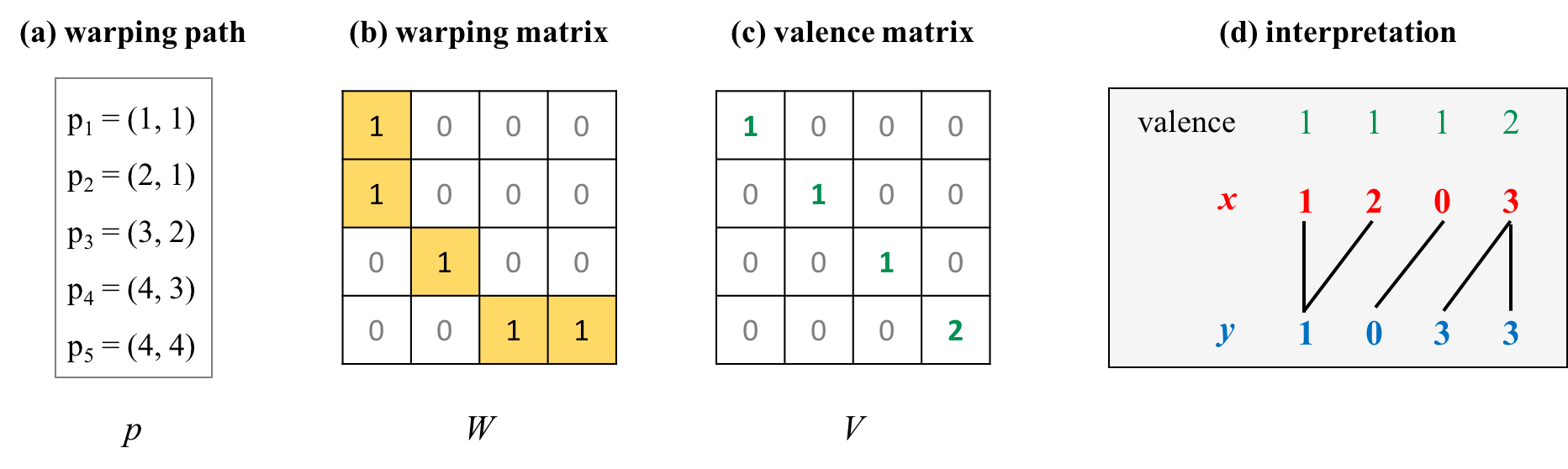}
\caption{Illustration of warping and valence matrix. Box (a) shows the warping path $w$ of Figure \ref{fig:wpath}. Box (b) shows the warping matrix $W$ of warping path $w$. The points $w_l = (i_l, j_l)$ of warping path $w$ determine the positions of the ones in the warping matrix $W$. Box (c) shows the valence matrix $V = (v_{ij})$ of warping path $w$. The matrix $V$ is a diagonal matrix, whose diagonal elements $v_{ii}$ are the row sums of $W$. Box (d) interprets the valence matrix $V$. The valence $v_{ii}$ of element $x_i$ is the number of elements in time series $y$ that are aligned to $x_i$ by warping path $w$. In other words, the valence $v_{ii}$ is the number of black lines emanating from element $x_i$.}
\label{fig:valence}
\end{figure}

To update prototype $p$, we pick a subgradient $g_{p,x} \in \partial D_x(p)$ as surrogate time series. The subgradients we use for updating prototypes can be expressed by warping and valence matrices \cite{Schultz2017}. 
\begin{definition}
Let $w \in \S{W}_{m,n}$ be a warping path. 
\begin{enumerate}
\item The \emph{warping matrix} of $w$ is a matrix $W \in \{0,1\}^{m \times n}$ with elements
\begin{equation*}
W_{ij} = \begin{cases} 
1 & (i,j) \in w \\ 
0 & \text{otherwise} 
\end{cases}.
\end{equation*}
\item The \emph{valence matrix} of $w$ is the diagonal matrix $V \in \N^{m \times m}$ with elements
\begin{align*}
V_{ii} = \sum_{j=1}^n W_{ij}.
\end{align*}
\end{enumerate}
\end{definition}
Figure \ref{fig:valence} provides an example of a warping and valence matrix. The warping matrix is a matrix representation of the corresponding warping path. The valence matrix is a diagonal matrix, whose elements count how often an element of the first time series is aligned to an element of the second one. The next definition introduces subgradients of squared DTW distortions.

\begin{proposition}[\cite{Schultz2017}]
Let $x \in \S{T}$ be a time series and $D_x: \R^m \rightarrow \R$ be a squared DTW distortion. Then the subdifferential $\partial D_x(p)$ contains a subgradient of the form
\begin{align*}
2 (Vp - Wx) \in \partial D_x(p),
\end{align*}
where $W$ and $V$ are the warping and valence matrix of an optimal warping path $w \in \S{W}(p,x)$. 
\end{proposition}
Using the concept of subgradients as surrogate for gradients, the generic scheme of LVQ in DTW spaces is as follows: 

\bigskip

\begin{samepage}
\hrule
\begin{enumerate}
\itemsep0em
\item Initialize codebook $\S{C}$. 
\item Repeat until termination:
\begin{enumerate}
\itemsep0em
\item Randomly select a training example $(x, y) \in \S{D}$.
\item For all prototypes $(p, z) \in \S{C}$ do
\begin{enumerate}
\item Compute an optimal warping path $w$ between $p$ and $x$.
\item Derive warping matrix $W$ and valence matrix $V$ of path $w$. 
\item Update prototype $p$ according to the rule
\begin{align}\label{eq:generic-DTW-update-rule}
p \;\leftarrow\; p - \eta \cdot f(p, x)\cdot(Vp - Wx),
\end{align}
where $\eta$ is an adaptive learning rate and $f$ is a class-compatible force function. 
\end{enumerate}
\item Optionally adjust hyper-parameters.
\end{enumerate}
\end{enumerate}
\hrule
\end{samepage}

\bigskip

Update rule \eqref{eq:generic-DTW-update-rule} is generic in the following sense: Every LVQ algorithm in Euclidean spaces whose update rule fits into the form of \eqref{eq:generic-Euclidean-update-rule} can be generalized to its corresponding counterpart in DTW spaces. In particular, LVQ1 and GLVQ are examples that can be directly generalized to DTW spaces.

In addition, the update rule \eqref{eq:generic-DTW-update-rule} in DTW spaces generalizes the update rule \eqref{eq:generic-Euclidean-update-rule} in Euclidean spaces in the following sense: Squared Euclidean distances between time series $p$ and $x$ of the same length $m$ correspond to the cost of aligning $p$ and $x$ along a warping path $w = (w_1, \ldots, w_m)$ with points $w_l = (l,l)$. Then the valence and warping matrix of $w$ are identity matrices and the subgradient of the squared DTW distortion reduces to the gradient of the squared Euclidean distortion.

\subsection{Stochastic Subgradient Update Rules}

This section illustrates that stochastic gradient descent rules for minimizing differentiable LVQ cost functions in Euclidean spaces extend to stochastic subgradient methods for minimizing their counterparts in DTW spaces. The following analysis is restricted to the cost function of GLVQ, but can be easily applied to other sufficiently well-behaved cost functions. By
\[
E_{\dtw} = \sum_{i=1}^N h(\kappa_{\dtw}(x_i)),
\]
we denote the cost function of asymmetric GLVQ in DTW spaces, where 
\[
\kappa_{\dtw}(x) = \frac{\delta^+ - \delta^-}{\delta^+ + \delta^-} 
\]
is the relative DTW distance difference at current input $x$. The distances $\delta^+$ and $\delta^-$ in $\kappa_{\dtw}(x)$ are the DTW distortions 
\begin{align*}
\delta^+ = \delta^2\args{p^+ - x} \qquad \text{and} \qquad \delta^- = \delta^2\args{p^- - x},
\end{align*}
The DTW distortion can be expressed as
\[
D_x(p) = \min_{p} C_p(p, x),
\]
where $C_p(p,x)$ is the cost of aligning $p$ and $x$ along warping path $p$. 

The relative difference $\kappa_{\dtw}(x)$ as a function of the prototypes $p^{\pm}$ has almost the same analytical properties as the relative difference $\kappa$ in Euclidean spaces:
\begin{enumerate}[(i)]
\itemsep0em
\item $\kappa_{\dtw}(x)$ is undefined if $\delta^+ = \delta^- = 0$.
\item $\kappa_{\dtw}(x)$ is locally Lipschitz continuous if $d^+ + d^- \neq 0$.
\item $\kappa_{\dtw}(x)$ is non-differentiable if one of the prototypes $p^{\pm}$ is not uniquely determined for $x$.
\item $\kappa_{\dtw}(x)$ can be non-differentiable if an optimal warping path between $x$ and $p^{\pm}$ is not unique.
\end{enumerate}
Case (i)--(iii) are as in the Euclidean space with a slight difference in case (i). The relative difference $\kappa$ in Euclidean spaces has a unique singularity for a given input $x$ due to the identity of indiscernibles satisfied by any metric (cf.~Section \ref{subsec:DTW}). Since the DTW distance does not satisfy the identity of indiscernibles, there are time series $p^+$, $p^-$ and $x$ that are pairwise distinct but nevertheless give $\delta^+ = \delta^- = 0$.

To show case (ii), observe that the cost function $C_p(p,x)$ is differentiable and therefore locally Lipschitz continuous \cite{Schultz2017}. Since local Lipschitz continuity is closed under the min-operation and rules of calculus, we find that $\kappa_{\dtw}(x)$ is also locally Lipschitz continuous, whenever $d^+ + d^- \neq 0$. Consequently and when well-defined, the function $\kappa_{\dtw}(x)$ is differentiable almost everywhere by Rademacher's Theorem \cite{Evans1992}. As in Euclidean spaces, we again have zero probability that non-differentiability occurs. The difference is that $\kappa_{\dtw}(x)$ has 'more' non-differentiable points than its counterpart in Euclidean spaces due to case (iv). 

For minimizing the cost function $E_{\dtw}$, the strategy to cope with singularities of case (i) and non-differentiabilities of case (iii) is as in the Euclidean space. We resolve non-differentiabilities of case (iv) by breaking ties and choosing one optimal warping path between $x$ and its closest prototypes $p^+$ and $p^-$. Apart from the singularities, the GLVQ algorithm in DTW spaces has the form of a stochastic subgradient method \cite{Bagirov2014,Schultz2017}.

\subsection{Relationship to The Symmetric LVQ1 Algorithm}\label{subsec:symmetric-asymmetric}

In 1998, Kohonen and Somervuo \cite{Kohonen1998} laid the foundation for extending (un)supervised prototype learning to arbitrary distance spaces. In a follow-up paper, Somervuo and Kohonen \cite{Somervuo1999} generalized the LVQ1 algorithm to DTW spaces. The technical difference between the Somervuo-Kohonen method and the proposed generic update rule \eqref{eq:generic-DTW-update-rule} is that the former implements symmetric and the latter asymmetric averaging.

To describe asymmetric and symmetric averaging, we first note that the generic Euclidean update rule can be written as a weighted average of the form
\[
p' = p - \alpha(p-x) = (1-\alpha)p - \alpha x,
\] 
where $x$ is the current input, $p$ is a prototype to be updated, $p'$ is the updated prototype, and $\alpha = \eta f(p,x)$ is a short-cut notation for the step size. The weighted average as update rule can be directly implemented in DTW spaces in an asymmetric and in a symmetric form. To describe both forms of an average, we assume that $w= (w_1, \ldots, w_L)$ is an optimal warping path between prototype $p$ and input $x$ with points $w_l =(i_l, j_l)$.

\medskip

\noindent
\emph{Asymmetric averaging.}
The warping path $w$ aligns every element $p_i$ of prototype $p$ with elements $x_{j+1} \ldots, x_{j+v}$ of input $x$, where $v \geq 1$ is the valency of $p_i$. Then the weighted asymmetric average $p'$ is given by
\[
p'_i =  (1-\alpha) \cdot v \cdot p_i + \alpha \cdot \args{x_{j_1} + \cdots + x_{j_q}}
\]
for all time points $i$ of prototype $p$. By construction, the length of $p'$ coincides with the length of $p$. The asymmetric weighted average corresponds to the generic update rule \eqref{eq:generic-DTW-update-rule} in element-wise rather than matrix notation. Asymmetric averages are asymmetric, because the length of a prototype is determined by choosing a reference time axis of predetermined and fixed length. 

\medskip

\noindent
\emph{Symmetric averaging.}
To average time series independent of the choice of a reference time axis, Kruskal and Liberman \cite{Kruskal1983} proposed symmetric averages. The weighted symmetric average $p'$ is of the form
\[
p'_l = (1-\alpha)p_{i_l} + \alpha x_{j_l}
\]
for all points $w_l = (i_l, j_l)$ of the optimal warping path $w$. The resulting prototype $p'$ has the same length $L$ as the warping path $w$. Repeated updating in symmetric form results in increasingly long prototypes. Somervuo and Kohonen applied an interpolation technique to adjust the length of the prototype $p'$. In doing so, the complete Somervuo-Kohonen update rule is defined by
\begin{align*}
p'_l = \Pi_m\Big((1-\alpha)p_{i_l} + \alpha x_{j_l}\Big)
\end{align*}
where $\Pi_m:\S{T} \rightarrow \S{T}_m$ denotes a (possibly non-linear) projection to the subset $\S{T}_m$. The symmetric update rule is generally not based on a subgradient of the distortion. In this sense, symmetric LVQ1 lacks a theoretical underpinning.

\section{Experiments}

The goal of these experiments is to assess the performance of asymmetric LVQ1 and GLVQ. 

\subsection{Data}

We used $30$ datasets from the UCR time series classification and clustering repository \cite{Chen2015}. The datasets were chosen to cover various characteristics such as application domain, length of time series, number of classes, and sample size (see Table \ref{tab:results}).  

\subsection{Algorithms}

\newcommand{\nn}{\texttt{1-nn}~}
\newcommand{\rglvq}{\texttt{rglvq}~}
\newcommand{\kmeans}{\texttt{kmeans}~}
\newcommand{\slvq}{\texttt{slvq}~}
\newcommand{\alvq}{\texttt{alvq}~}
\newcommand{\glvq}{\texttt{glvq}~}

We considered the following nearest neighbor classifiers:
\begin{center}
\begin{tabular}{l@{\qquad}l@{\qquad}c}
\hline
Notation & Algorithm & Reference\\
\hline
\\[-2ex]
\nn    		& nearest neighbor classifier 	& traditional\\
\rglvq		& relational generalized LVQ 	& \cite{Gisbrecht2012}\\
\kmeans 		& k-means classifier 			& \cite{Petitjean2014,Petitjean2016}\\
\slvq  		& symmetric LVQ1 				& \cite{Somervuo1999} \\
\alvq   		& asymmetric LVQ1 				& proposed\\
\glvq    	& asymmetric generalized LVQ 	& proposed\\
\hline
\end{tabular}
\end{center}
The codebook of the \nn classifier consists of the entire training set. All other algorithms use one prototype per class to obtain fast NN classifiers. The \rglvq classifier assumes as input a matrix of pairwise DTW dissimilarities. Then \rglvq learns prototypes in the dissimilarity space (without Nystr\"om approximation) \cite{Gisbrecht2012}. The \kmeans classifier refers to the nearest neighbor classifier whose prototypes were learned by using the unsupervised k-means algorithm together with the DBA algorithm for recomputing the centroids \cite{Petitjean2014,Petitjean2016}. The \slvq method is the symmetric Somervuo-Kohonen extension of LVQ1 to DTW spaces \cite{Somervuo1999}.

\subsection{Experimental Protocol}

To assess the performance of the nearest prototype classifiers, we conducted ten-fold cross validation and reported the average classification accuracy, briefly called accuracy, henceforth. We preferred cross-validation over hold-out validation via the supplied train-test splits for two reasons: first, to obtain better estimates of the generalization performance; and second, to make sense of data reduction by increasing the size of the training set to $90\%$ of all data. 

For the \rglvq classifier, we used the default setting as provided by the publicly available matlab-implementation.\footnote{\url{https://www.techfak.uni-bielefeld.de/~fschleif/software.xhtml} (January 2017).} The default-setting has also been used in experiments and is justified, because \rglvq is not sensitive to its hyper-parameters \cite{Mokbel2015}. 

The \kmeans classifier applied the k-means clustering method to every class separately. Since the codebook of \kmeans consists of one prototype per class, k-means clustering reduced to averaging warped time series of the same class. For time series averaging, we applied the DBA algorithm \cite{Petitjean2011}. The DBA algorithm terminated latest after $50$ iterations. 

All non-relational LVQ variants were intialized by the centroids of the \kmeans classifier and terminated latest after $1,000$ iterations. The hyper-parameters of the LVQ algorithms were selected from
\begin{align*}
\sigma &\in \cbrace{0.1, 0.5, 1, 5, 10, 25, 50} \\
\eta &\in \cbrace{0.001, 0.0025, 0.005, 0.01, 0.025, 0.05, 0.1, 0.25, 0.5},
\end{align*}
where $\sigma$ is the initial slope parameter of \glvq and $\eta$ is the learning-rate of \slvq and \texttt{alvq}. The learning rate of \glvq was set to one. The slope parameter was adjusted by $\sigma \cdot t$, where $t$ is the number of epochs. The respective hyper-parameter of a given LVQ algorithm \texttt{A} was selected for the $i$-th fold $\overline{\S{D}}_i$ according to the following procedure: (1) Train \texttt{A} with every hyper-parameter value on the $i$-th training set $\S{D}_i$. (2) Select the trained model with the lowest error rate on training set $\S{D}_i$. (3) Test the chosen model on the $i$-th fold $\overline{\S{D}}_i$.

\subsection{Results}

\definecolor{crk01}{rgb}{0.25, 1.0, 0}
\definecolor{crk02}{rgb}{1.0, 1.0, 0}
\definecolor{crk03}{rgb}{1.0, 0.75, 0}
\definecolor{crk04}{rgb}{1.0, 0.1, 0}
\newcommand{\cI}{\cellcolor{crk01}}
\newcommand{\cII}{\cellcolor{crk02}}
\newcommand{\cIII}{\cellcolor{crk03}}
\newcommand{\cIV}{\cellcolor{crk04}}

\begin{table}
\centering
\begin{tabular}{lr@{\qquad}rrrrrr}
\hline
\hline
Dataset & $N/K$ & \nn & \rglvq & \kmeans & \slvq & \alvq & \glvq \\
\hline
\\[-2ex]
Beef & 12.0 & \cII 53.3 & \cIII 41.7 & \cIII 43.3 & \cIV 40.0 & \cIII 51.7 & \cI 63.3\\
CBF & 310.0 & \cI 99.9 & \cIII 93.9 & \cIII 96.6 & \cIV 93.6 & \cIII 96.1 & \cII 99.6\\
ChlorineConcentration & 1435.7 & \cI 99.6 & \cIII 45.5 & \cIV 34.9 & \cIII 55.0 & \cIII 56.2 & \cII 71.6\\
Coffee & 28.0 & \cI 100.0 & \cIII 96.7 & \cIII 96.4 & \cIII 96.4 & \cIII 96.4 & \cII 98.2\\
ECG200 & 100.0 & \cI 83.5 & \cIII 72.0 & \cIII 71.0 & \cIII 71.0 & \cIII 74.5 & \cII 80.5\\
ECG5000 & 1000.0 & \cII 93.3 & \cIV 80.8 & \cIII 83.1 & \cIII 89.4 & \cIII 91.1 & \cI 94.3\\
ECGFiveDays & 442.0 & \cI 99.2 & \cIII 63.9 & \cIV 63.0 & \cIII 72.7 & \cIII 71.0 & \cII 99.1\\
ElectricDevices & 2376.7 & \cI 79.2 & \cIII 63.5 & \cIV 56.5 & \cIII 57.4 & \cIII 64.8 & \cII 78.2\\
FaceFour & 28.0 & \cI 92.9 & \cIII 88.4 & \cIV 86.6 & \cIII 87.5 & \cIII 87.5 & \cII 92.0\\
FacesUCR & 160.7 & \cI 97.8 & \cIV 82.2 & \cIII 85.6 & \cIII 83.3 & \cIII 85.3 & \cII 97.2\\
Fish & 50.0 & \cII 80.3 & \cIII 63.4 & \cIII 65.1 & \cIV 63.1 & \cIII 64.9 & \cI 90.9\\
Gun Point & 100.0 & \cII 91.5 & \cIV 46.0 & \cIII 66.0 & \cIII 66.5 & \cIII 69.0 & \cI 97.0\\
Ham & 107.0 & \cII 72.4 & \cIII 61.0 & \cIII 65.0 & \cIV 60.3 & \cIII 65.0 & \cI 74.8\\
ItalyPowerDemand & 548.0 & \cI 95.8 & \cIII 82.6 & \cIII 85.0 & \cIV 77.0 & \cIII 86.4 & \cII 95.3\\
Lighting2 & 60.5 & \cI 89.3 & \cIV 57.1 & \cIII 64.5 & \cII 76.9 & \cIII 62.0 & \cIII 76.0\\
Lighting7 & 20.4 & \cIII 71.3 & \cII 79.1 & \cIII 76.2 & \cIV 59.4 & \cIII 79.0 & \cI 82.5\\
MedicalImages & 114.1 & \cI 80.7 & \cIII 43.5 & \cIV 42.8 & \cIII 53.5 & \cIII 60.6 & \cII 71.7\\
OliveOil & 15.0 & \cI 85.0 & \cIV 81.7 & \cIII 83.3 & \cI 85.0 & \cIII 83.3 & \cI 85.0\\
ProximalPhalanxOutlineAgeGroup & 201.7 & \cIV 75.5 & \cI 85.6 & \cIII 81.3 & \cIII 82.2 & \cIII 82.6 & \cII 83.6\\
ProximalPhalanxOutlineCorrect & 445.5 & \cII 82.0 & \cIV 61.1 & \cIII 63.2 & \cIII 71.9 & \cIII 72.3 & \cI 85.1\\
ProximalPhalanxTW & 100.8 & \cII 77.5 & \cIII 76.7 & \cIV 72.2 & \cIII 76.4 & \cIII 75.9 & \cI 81.2\\
RefrigerationDevices & 250.0 & \cII 60.7 & \cIII 56.7 & \cIV 54.4 & \cIII 55.3 & \cIII 56.7 & \cI 62.9\\
Strawberry & 491.5 & \cI 96.5 & \cIII 60.3 & \cIV 58.8 & \cIII 74.7 & \cIII 78.8 & \cII 94.1\\
SwedishLeaf & 75.0 & \cII 82.0 & \cIII 68.5 & \cIII 69.2 & \cIII 69.6 & \cIV 68.4 & \cI 87.6\\
synthetic control & 100.0 & \cIII 99.2 & \cIV 93.5 & \cII 99.3 & \cIII 98.5 & \cIII 99.2 & \cI 99.5\\
ToeSegmentation1 & 134.0 & \cII 85.8 & \cIII 68.7 & \cIII 73.1 & \cIV 65.7 & \cIII 75.8 & \cI 92.5\\
Trace & 50.0 & \cI 100.0 & \cIV 95.5 & \cIII 99.0 & \cI 100.0 & \cIII 99.5 & \cI 100.0\\
Two Patterns & 1250.0 & \cI 100.0 & \cIII 99.8 & \cIII 97.9 & \cIV 96.9 & \cIII 98.1 & \cII 99.9\\
Wafer & 3582.0 & \cI 99.4 & \cIII 63.2 & \cIV 45.0 & \cIII 89.6 & \cIII 91.7 & \cII 96.5\\
Yoga & 1650.0 & \cI 93.9 & \cIII 56.0 & \cIII 58.5 & \cIV 55.0 & \cIII 66.0 & \cII 73.2\\
\hline
\hline
\multicolumn{7}{r}{\footnotesize green: rank 1 --- yellow: rank 2 --- orange: rank 3, 4, 5 --- red: rank 6 }
\end{tabular}
\caption{Average accuracies of six nearest neighbor classifiers on $30$ datasets using $10$-fold cross validation. The second column $N/K$ roughly shows the average number of elements per class.}
\label{tab:results}
\end{table}

\begin{table}
\centering
\begin{tabular}{l@{\qquad}rrrrrr@{\qquad}cc}
\hline
\hline
Classifier & \cI rank 1 & \cII rank 2 & \cIII rank 3 & \cIII rank 4 & \cIII rank 5 & \cIV rank 6 & avg & std\\
\hline
\\[-2ex]
\nn 		& \cI 17 & \cII 10 & \cIII 1  & \cIII 0  & \cIII  1 & \cIV 1 & 1.70 & 1.18\\
\rglvq 	& \cI 1 	 & \cII 1  & \cIII 5  & \cIII 2  & \cIII 13 & \cIV 8 & 4.63 & 1.33\\
\kmeans 	& \cI 0  & \cII 1  & \cIII 4  & \cIII 9  & \cIII  7 & \cIV 9 & 4.63 & 1.16\\
\slvq  	& \cI 2  & \cII 1  & \cIII 2  & \cIII 11 & \cIII  5 & \cIV 9 & 4.43 & 1.43\\
\alvq 	& \cI 0  & \cII 0  & \cIII 18 & \cIII 9  & \cIII  2 & \cIV 1 & 3.53 & 0.78\\
\glvq 	& \cI 14 & \cII 15 & \cIII 1  & \cIII 0  & \cIII  0 & \cIV 0 & 1.57 & 0.57\\
\hline
\hline
\end{tabular}
\caption{Rank distribution, average ranks, and standard deviation. The average accuracy of every classifier on a given dataset was ranked, where ranks go from $1$ (highest accuracy) to $6$ (lowest accuracy).}
\label{tab:ranks}
\end{table}

This section presents and discusses the results. Table \ref{tab:results} summarizes the $10$-fold cross validation results of the six classifiers on $30$ datasets. 

\subsubsection*{General Performance Comparison}

To assess the generalization performance of the six classifiers, we compared their average classification accuracies.  
In an overall comparison, \glvq was ranked first with average rank $1.57$ closely followed by $\nn$ with average rank $1.70$. The other four classifiers were left behind with nearly two- to three-rank gaps (Table \ref{tab:ranks}). The asymmetric LVQ variants \glvq and \alvq exhibited the most stable rankings with standard deviation of $0.57$ and $0.78$, resp., whereas the symmetric Somervuo-Kohonen method \slvq was most unstable with standard deviation $1.43$.

\begin{figure}[t]
\centering
\begin{tabular}{cc}
\textbf{Winning Percentage} & \textbf{Mean Percentage Difference}\\[1ex]
\includegraphics[width=0.47\textwidth]{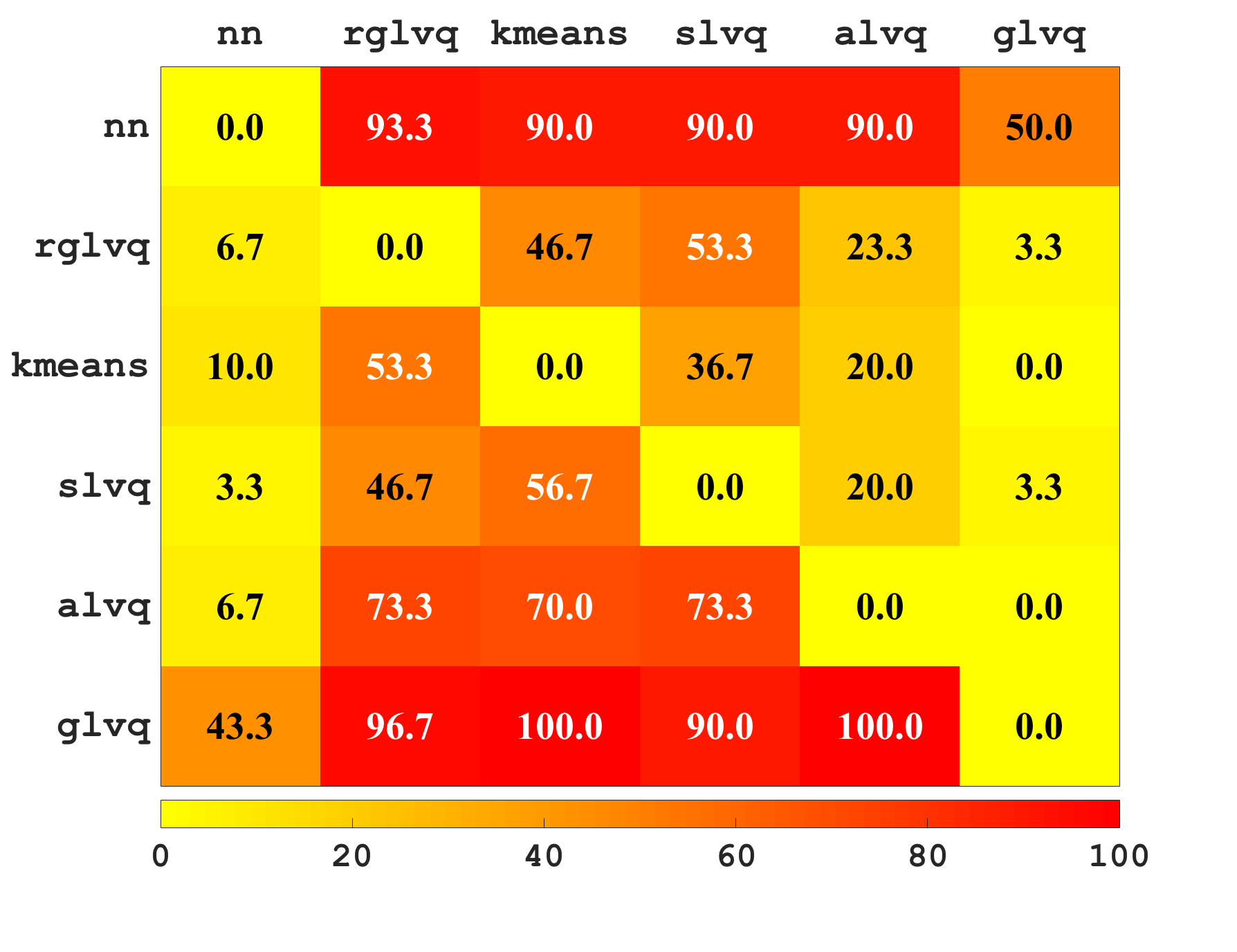} &
\includegraphics[width=0.47\textwidth]{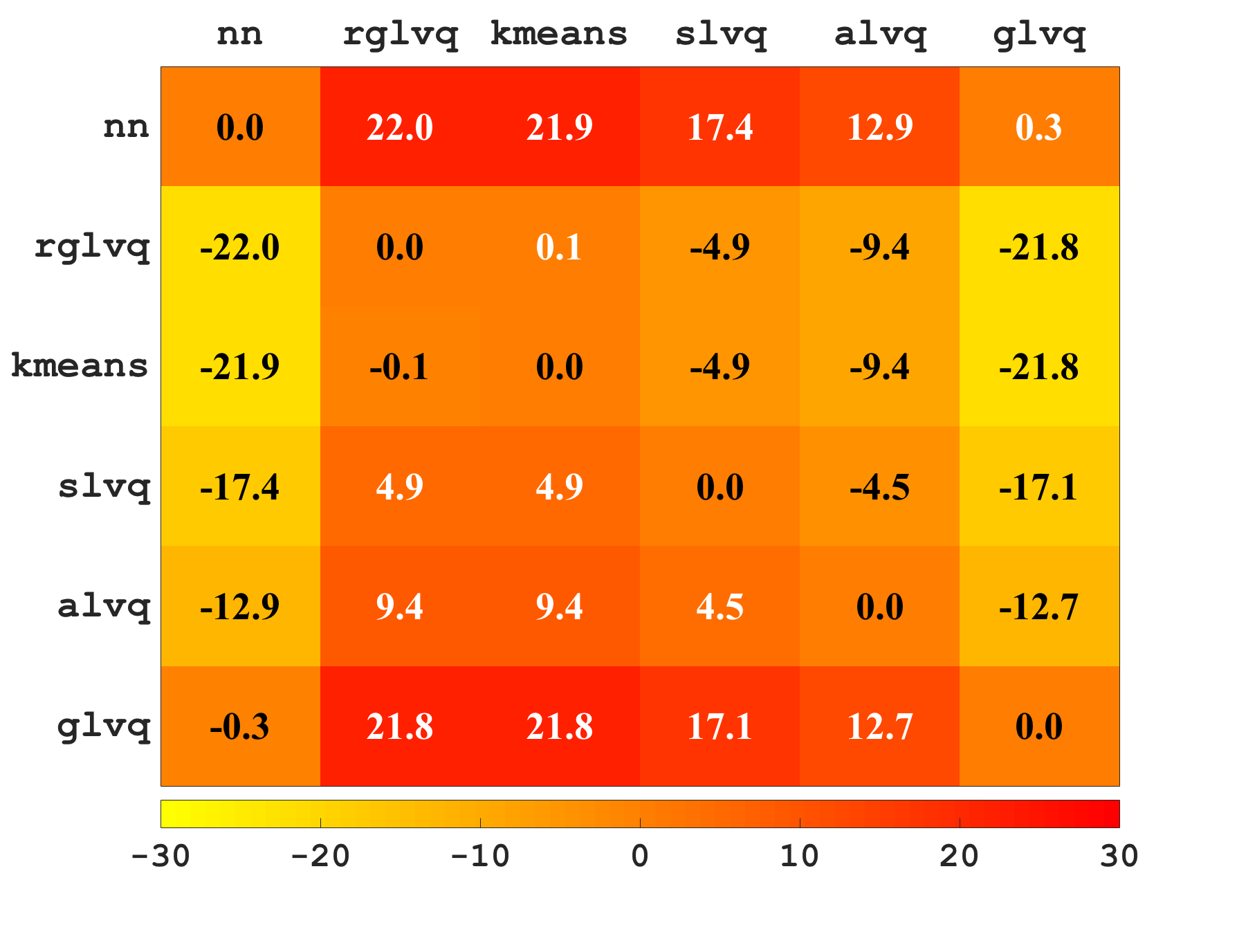}
\end{tabular}
\caption{Pairwise comparison of results. Left: Pairwise winning percentages $w_{ij}$, where classifier in row $i$ wins $w_{ij}$ percentages of all competitions against the classifier in column $j$. Right: Pairwise mean percentage difference $a_{ij}$ in accuracy, where the accuracy of classifier in row $i$ is $a_{ij}$ percentages better on average than the accuracy of the classifier in column $j$. A definition of both measures is given in Appendix \ref{sec:performance-measures}.} 
\label{fig:rnk}
\end{figure}

To assess the differences in accuracy between the six classifiers, we compared the results of Table \ref{tab:results} in a pairwise manner. The mean percentage difference in accuracy between the two best performing classifiers (\texttt{1-nn}, \texttt{glvq}) and the other four classifiers (\texttt{rglvq}, \texttt{kmeans}, \texttt{slvq}, \texttt{alvq}) ranged from $12.7 \%$ to $22.0 \%$ (Figure \ref{fig:rnk}, right panel). These results show that the accuracies of the two best performing classifiers were substantially higher than the accuracies of the other four classifiers by a large margin. 

A pairwise comparison of \nn and \glvq exhibited slight advantages for the \nn classifier. The \nn classifier won more often than the \glvq classifier ($50 \%$ vs.~$43.4 \%$) but differences in classification accuracy were marginal ($0.3 \%$) on average (Figure \ref{fig:rnk}). The \nn and \glvq classifier complement each other in the sense that each of both classifiers substantially outperformed the other one on six datasets by more than $10$ percentage points (Table \ref{tab:results}): 
\begin{enumerate}
\itemsep0em
\item \nn outperformed \glvq on \texttt{ChlorineConcentration}, \texttt{Lighting2}, and \texttt{Yoga}. 
\item \glvq outperformed \nn on \texttt{Beef}, \texttt{Fish}, and \texttt{Lighting7}.
\end{enumerate}
The first item suggests that one prototype per class is insufficient for separating the classes of \texttt{Chlorine\-Con\-centration}, \texttt{Lighting2}, and \texttt{Yoga}. The second item indicates that the \nn classifier suffers from noisy training examples that degrade its generalization performance on \texttt{Beef}, \texttt{Fish}, and \texttt{Lighting7}.

With regard to computation time, the four prototype generation methods \texttt{kmeans}, \texttt{slvq}, \texttt{alvq}, and \texttt{glvq} are substantially faster than the \nn and \rglvq classifier. When classifying a test example, computation time is dominated by the number of DTW distance calculations. By construction, the four prototype generation methods computed $K$ distances, where $K$ is the number of classes. In contrast, the \nn and the \rglvq classifier required $N$ distance computations, where $N$ is the number of training examples. In this study, the average speed-up factor $N/K$ of the four prototype generation methods over \nn and \rglvq ranged from one to three orders of magnitude (see Table \ref{tab:results} and compute $0.9\cdot N/K$ due to 10-fold cross-validation).

The results indicate that \glvq best traded solution quality against computation time. The solution quality of \glvq and \nn were comparable, whereas the solution quality of \texttt{rglvq}, \texttt{kmeans}, \texttt{slvq}, and \texttt{alvq} were not competitive. These findings suggest that \glvq is a suitable alternative in online settings and situations, where storage is limited and short classification times are required. 

\subsubsection*{Unsupvervised vs.~Supervised}

Compared to other prototype selection and generation methods, the \kmeans classifier showed convincing results in time series classification \cite{Petitjean2014,Petitjean2016} using a pre-specified train-test split supplied by the contributors of the datasets. To assess the effect of supervised prototype adaption, we compared the classification accuracies of the \kmeans classifier and the three LVQ classifiers in DTW space (\texttt{slvq}, \texttt{alvq}, \texttt{glvq}). 

The \glvq classifier won all competitions against \kmeans with $21.8$ mean percentage difference in classification accuracy. The other two LVQ methods (\texttt{slvq}, \texttt{alvq}) won more competitions than the \kmeans classifier with $4.9$ and $9.4$ mean percentage differences in accuracy, respectively (Figure \ref{fig:rnk}). These results suggest that LVQ methods are -- on average -- more suitable prototype generation methods for constructing memory- and time-efficient nearest neighbor classifiers than the \kmeans classifier. This finding is in line with similar results in Euclidean spaces \cite{Hastie2009}. 

\subsubsection*{Relational vs.~Non-Relational}

To assess the effects of relational vs.~non-relational prototype learning, we compared the results of \rglvq against the prototype generation methods \texttt{kmeans}, \texttt{slvq}, \texttt{alvq}, and \texttt{glvq}.

We made the following observations (Table \ref{tab:ranks} and Figure \ref{fig:rnk}): (i) the \rglvq classifier was ranked last together with \kmeans  and both classifiers have comparable generalization performance; (ii) all non-relational LVQ performed better than \texttt{rglvq}; (iii) the non-relational \glvq classifier has substantially higher accuracy than its relational counterpart \rglvq with mean percentage difference of almost $22 \%$; (iv) all non-relational classifiers were one to three orders of magnitude faster than \rglvq (cf.~disscussion on general performance comparison). 

The \rglvq classifier is neither competitive with \nn and also seems to be less suited for improving storage and computational requirements of the naive NN method. In all cases, \rglvq retained the entire training set in order to embed a new test example into the pseudo-Euclidean space. Techniques such as the Nystr\"om approximation can considerably improve storage and computation requirement but may compromise solution quality \cite{Gisbrecht2012,Mokbel2015}. These findings suggest that prototype learning in DTW spaces should be preferred over relational prototype learning in a pseudo-Euclidean space.

\subsubsection*{Symmetric LVQ1 vs.~Asymmetric LVQ1}

To assess the effects of the different types of update rules on the generalization performance, we compared the classification accuracies of \slvq against \texttt{alvq}.

In an overall comparison, the average rank of \alvq was almost one rank better than the average rank of \slvq ($3.53$ vs.~$4.43$, Table \ref{tab:ranks}). The rankings of \alvq were more stable than the rankings of \texttt{slvq}, which was the most unstable classifier (std $0.78$ vs.~std $1.43$, Table \ref{tab:ranks}). Consequently, \alvq was ranked last only once, whereas \slvq was ranked last $9$ times. Finally, in a pairwise comparison, \alvq won $73.3 \%$ of all competitions against \texttt{slvq} with $4.5$ mean percentage differences in accuracy (Figure \ref{fig:rnk}).

These results indicate that the asymmetric update rule of LVQ1 gives better and more stable results on average than its symmetric counterpart. These findings are in line with similar results on estimating the sample variance of a set of time series using symmetric and asymmetric averaging of time series in DTW spaces \cite{Petitjean2011,Soheily-Khah2015}. As pointed out in \cite{Schultz2017}, the reason for unstable and less accurate results of \slvq can be due to the projection step that may displace centroids in an uncontrolled manner.
Nevertheless symmetric averaging can but need not have negative effects as shown by the results of \slvq obtained on \texttt{Lighting2}, \texttt{OliveOil}, and \texttt{Trace}.

\subsubsection*{Varying the Number of Prototypes}

\begin{figure}[t]
\centering
\begin{tabular}{cc}
\includegraphics[width=0.45\textwidth]{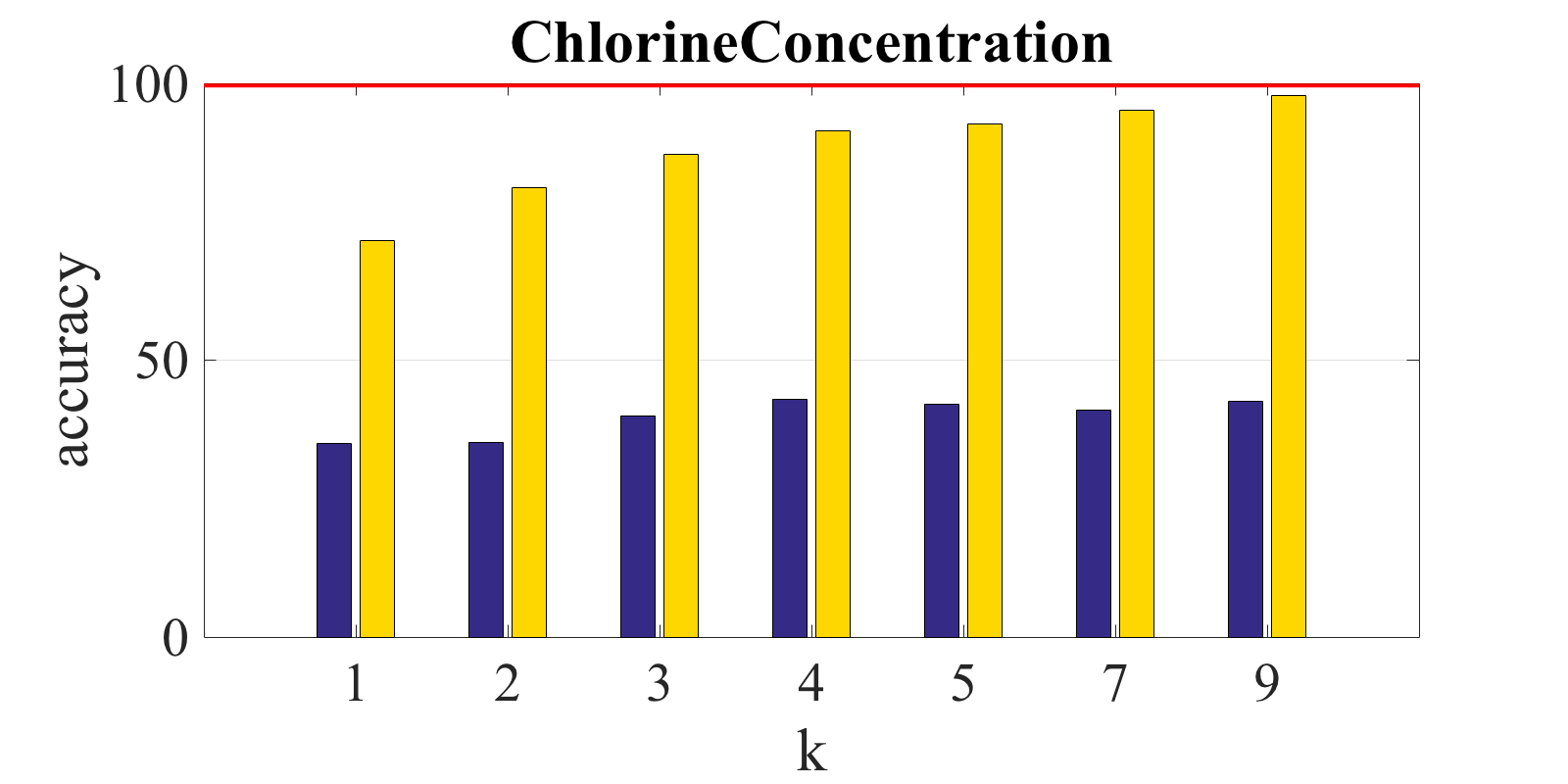} &
\includegraphics[width=0.45\textwidth]{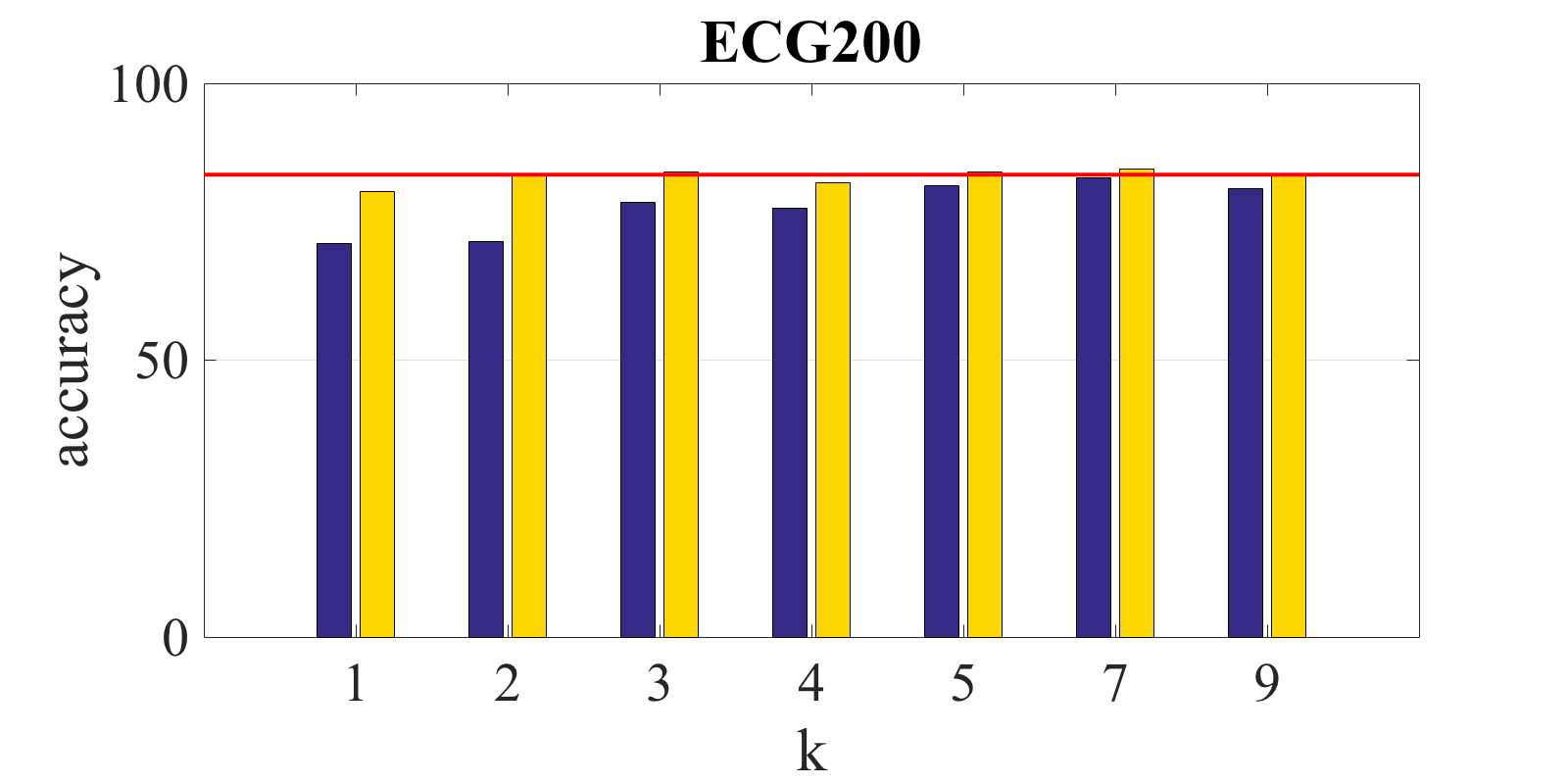}\\
\includegraphics[width=0.45\textwidth]{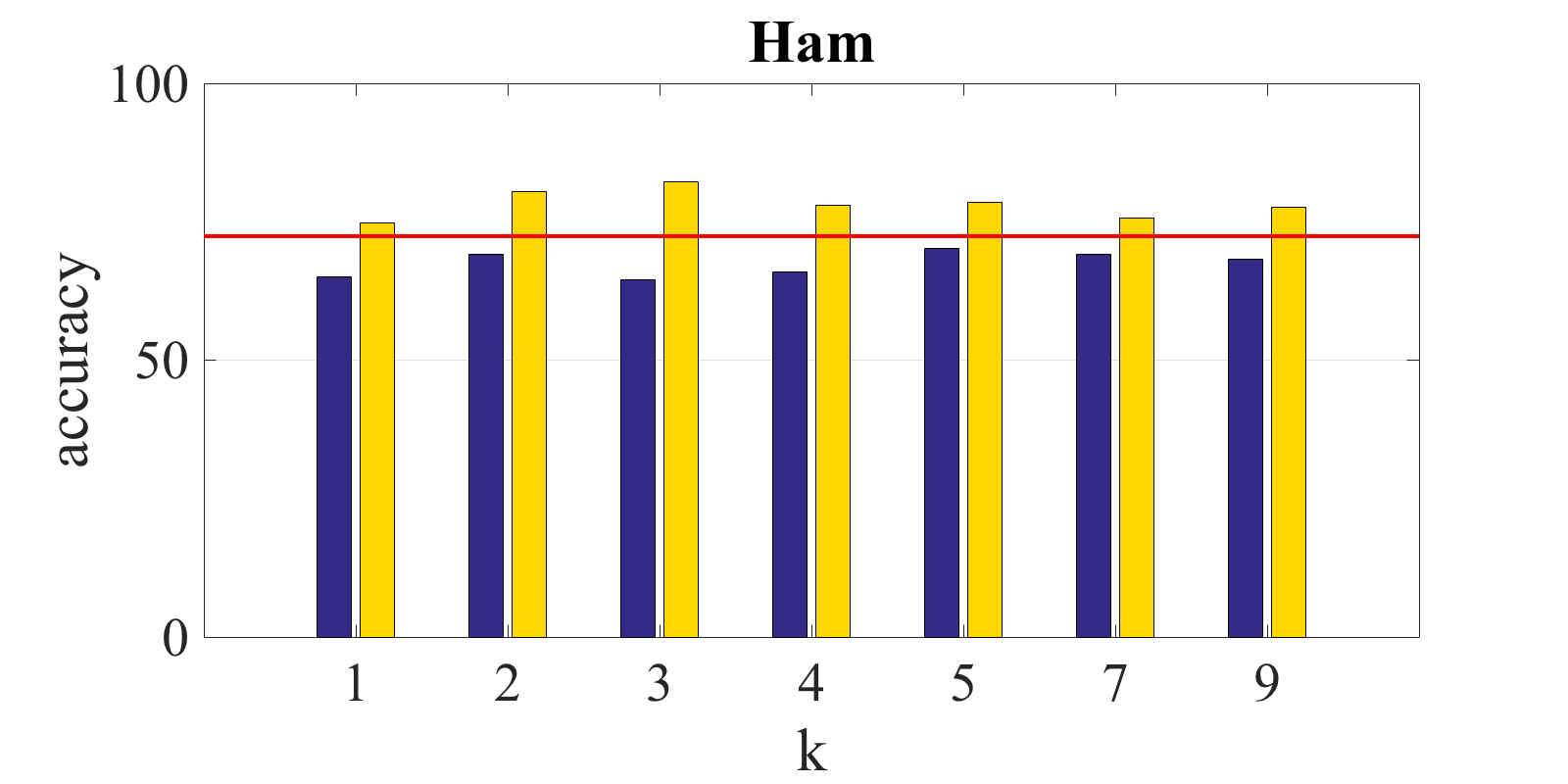} &
\includegraphics[width=0.45\textwidth]{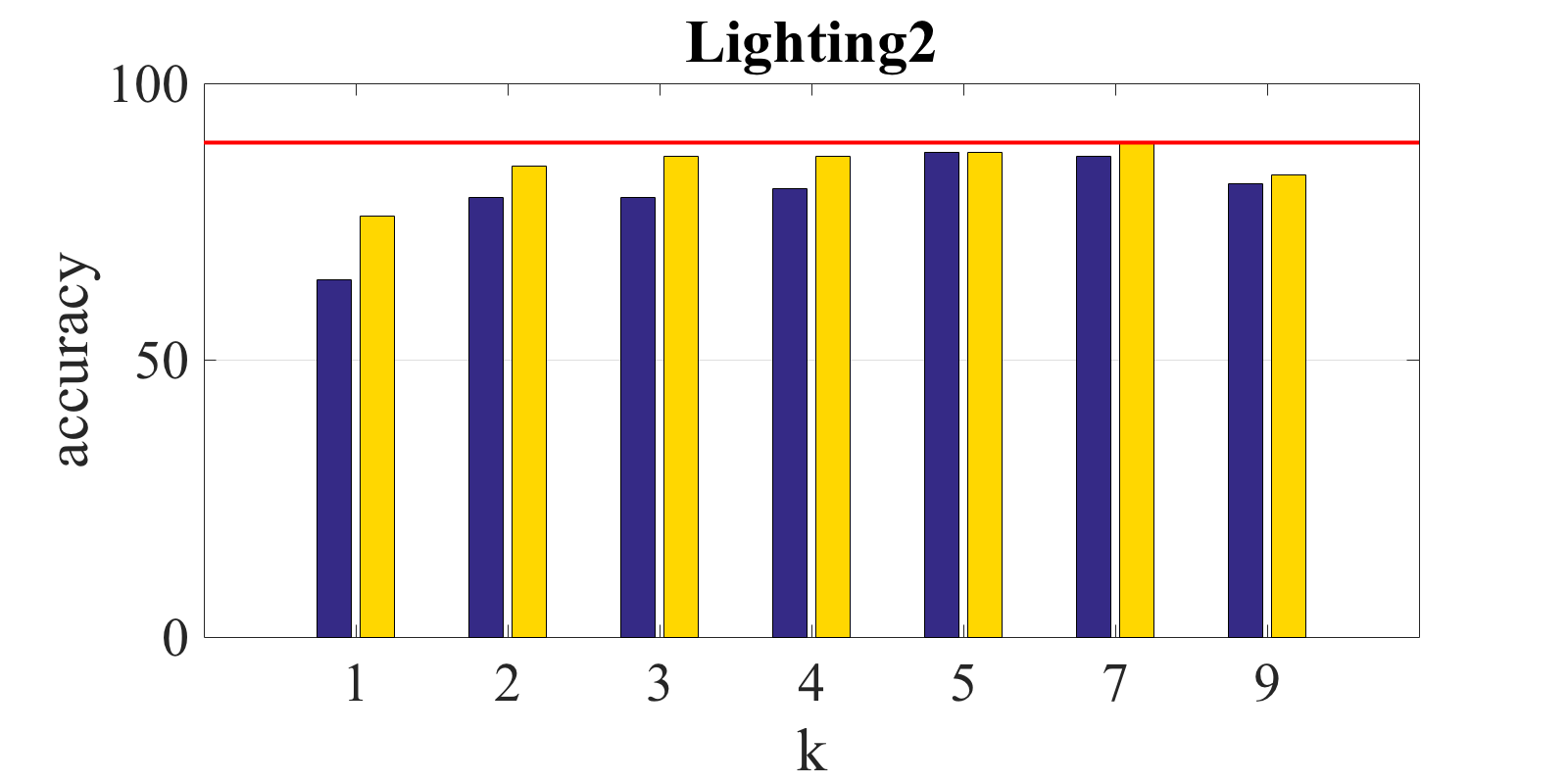}\\
\includegraphics[width=0.45\textwidth]{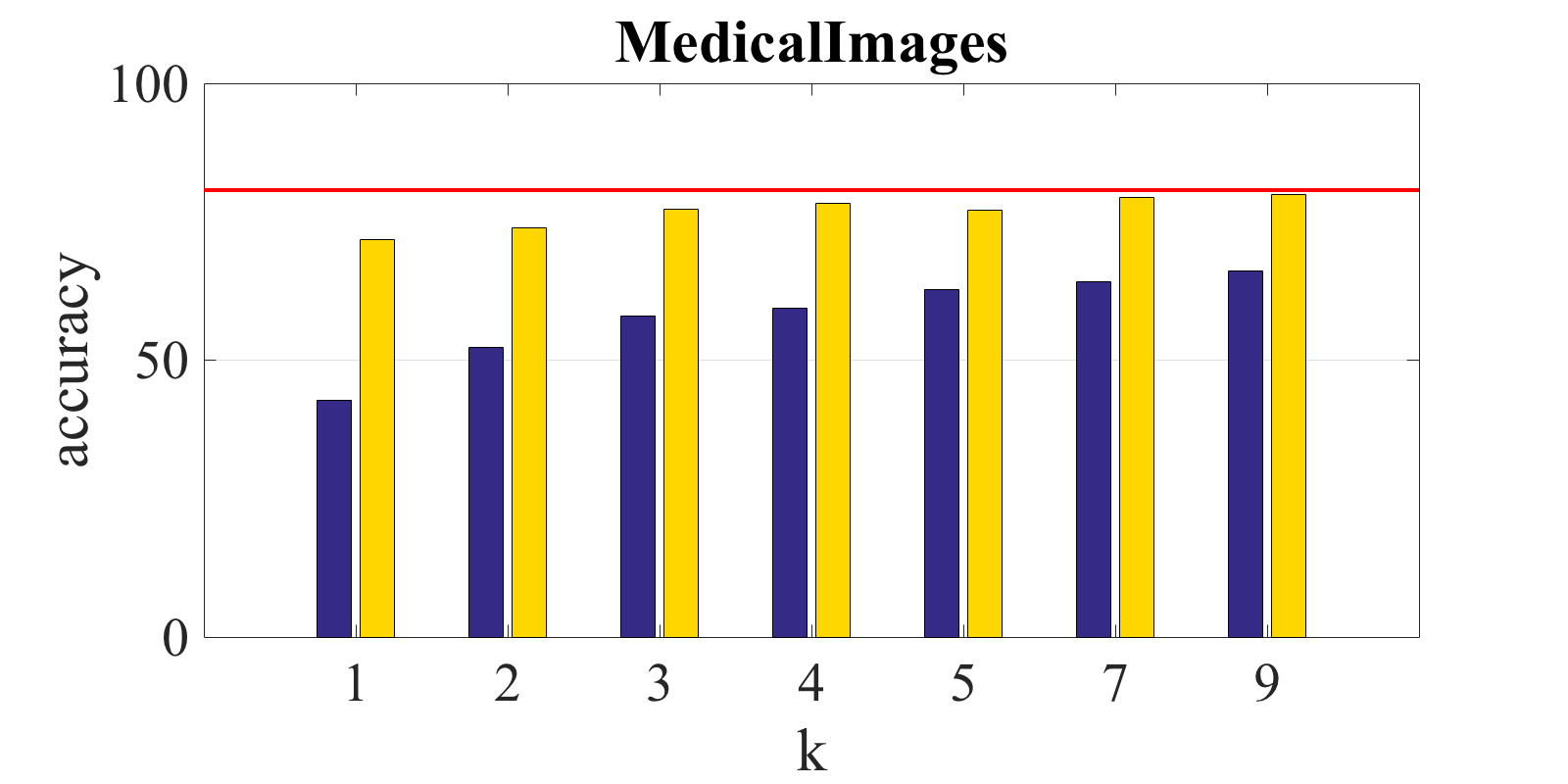} &
\includegraphics[width=0.45\textwidth]{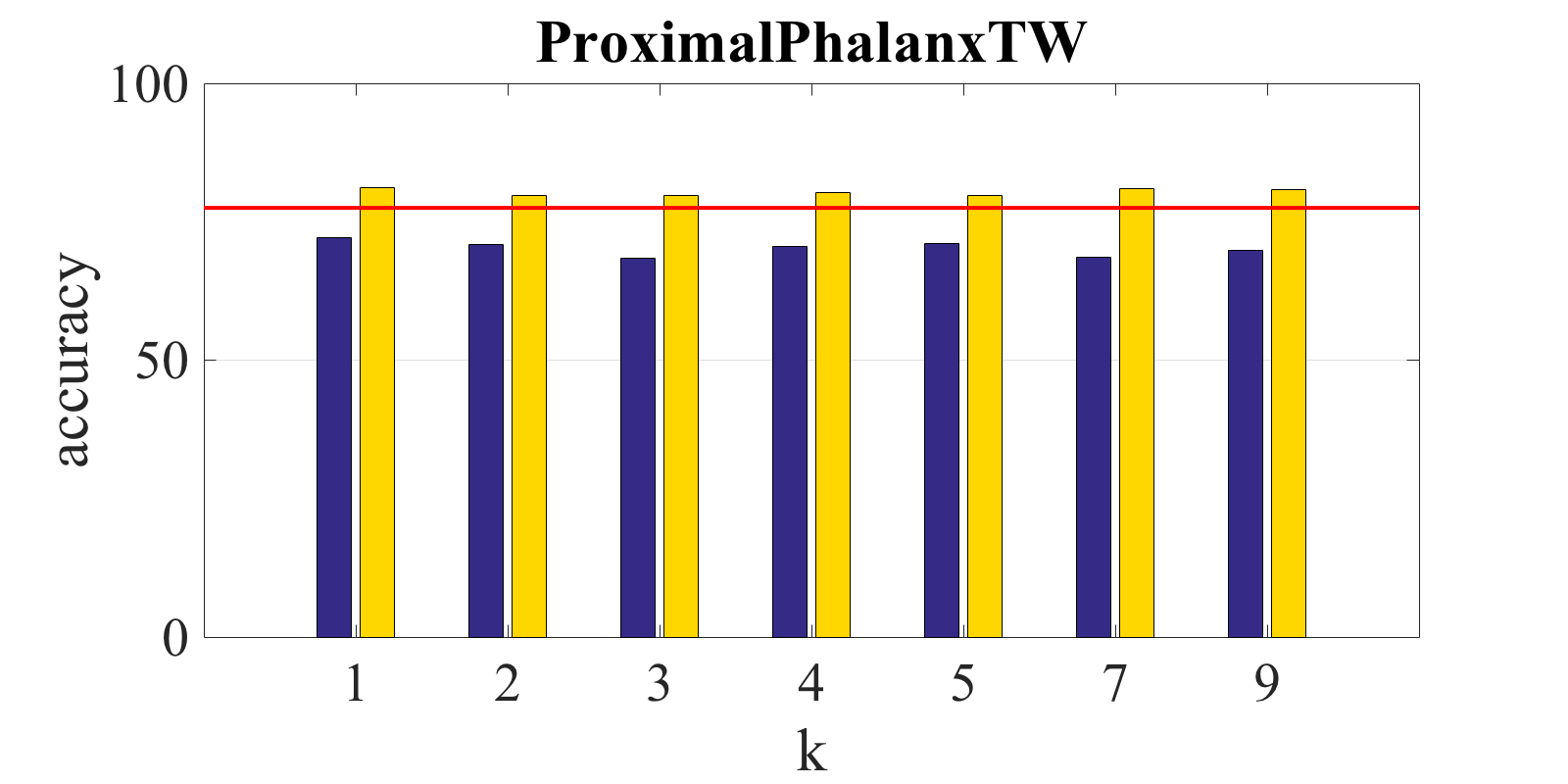}\\
\includegraphics[width=0.45\textwidth]{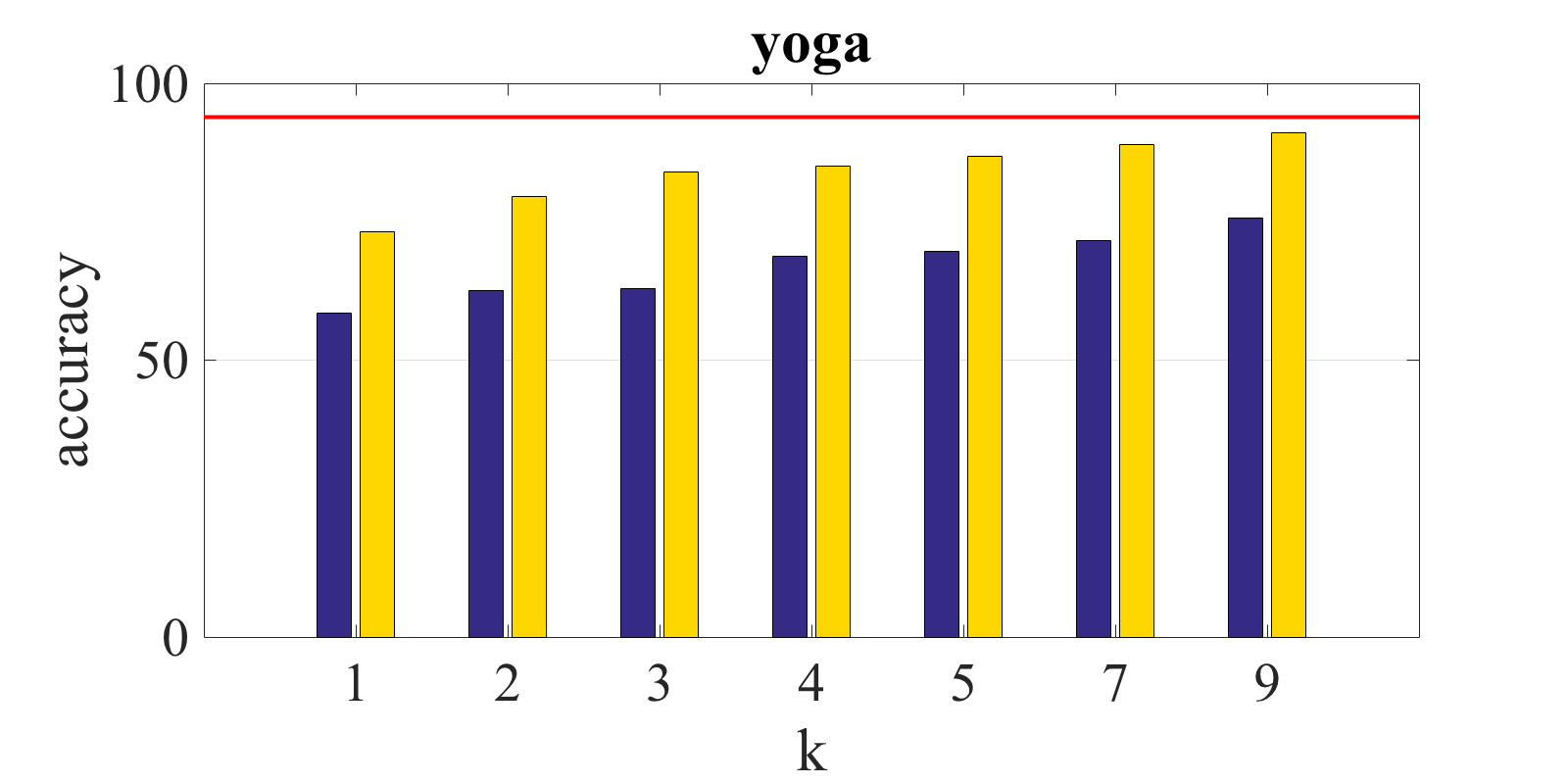} &
\raisebox{0.6cm}[1mm][1mm]{\includegraphics[width=0.3\textwidth]{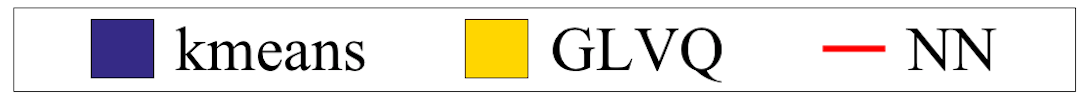}}\\
\end{tabular}
\caption{Accuracy of $10$-fold cross validation as a function of the number $k$ of prototypes per class.} 
\label{fig:res_k}
\end{figure}

To compare the dependence of the classification accuracy of \glvq and \kmeans on the number $k$ of prototypes per class, we considered seven difficult datasets for which the error rate of \glvq was higher than $10\%$ for $k = 1$. We used the same experimental protocol as before. Figure \ref{fig:res_k} summarizes the results.

The main observation is that a careful selection of a small number $k$ of prototypes per class was sufficient for \glvq to approach or even outperform the classification accuracy of the \nn classifier. We could not confirm this finding for the \kmeans classifier. The second observation is that \glvq performed consistently better than \kmeans for all $k$ and all seven datasets. These findings suggest that \glvq is a substantially more efficient alternative to the \nn classifier and is capable to maintain or even surpass the high classification accuracy often achieved by \texttt{1-nn}.

\section{Conclusion}
In this article, we presented a generic scheme to extend LVQ methods to DTW spaces. The generic update rule applies asymmetric averaging of warped time series. The asymmetric rule implements stochastic subgradient updating, which places the proposed LVQ scheme on an analytical foundation. Empirical results show that asymmetric GLVQ performed best compared to other state-of-the-art prototype generation methods by a large margin. In addition, we found that asymmetric LVQ1 is more stable and has higher generalization performance than its symmetric counterpart. As in Euclidean spaces, the non-relational supervised prototype generation methods outperformed the unsupervised k-means algorithm. The generalization performance of relational GLVQ is comparable with k-means. The results suggest that asymmetric GLVQ is a strong candidate for nearest neighbor classification in online settings and in situations where computation time and storage demands are an issue. The generic LVQ update rule can serve as a blueprint for directly extending unsupervised prototype learning methods to DTW spaces in a principled way, such as vector quantization, self-organizing maps, and neural gas. Finally, future work aims at studying the theoretical properties of asymmetric LVQ in DTW spaces.

\paragraph*{\textbf{Acknowledgements}.} B.~Jain was funded by the DFG Sachbeihilfe \texttt{JA 2109/4-1}.

\begin{small}

\appendix

\section{Performance Measures}\label{sec:performance-measures}

This section describes the pairwise winning percentage and pairwise mean percentage difference. 

\subsection{Winning Percentage}

The pairwise winning percentages are summarized in a matrix $W = (w_{ij})$. The winning percentage $w_{ij}$ is the fraction of datasets for which the accuracy of the classifier in row $i$ is strictly higher than the accuracy of the classifier in column $j$. Formally, the winning percentage $w_{ij}$ is defined by
\[
w_{ij} = 100 \cdot \frac{\abs{\cbrace{d \in \S{D} \,:\, \text{acc}_d(j) < \text{acc}_d(i)}}}{\abs{\S{D}}}
\]
where $\text{acc}_d(i)$ is the accuracy of the classifier in row $i$ on dataset $d$, and $\text{acc}_d(j)$ is the accuracy of the classifier in column $j$ on $d$. The percentage $w_{ij}^{eq}$ of ties between classifiers $i$ and $j$ can be inferred by 
\[
w_{ij}^{eq} = 100-w_{ij}-w_{ji}.
\]

\subsection{Pairwise Mean Percentage Difference}
The pairwise mean percentage differences are summarized in a matrix $A = (a_{ij})$. The mean percentage difference $a_{ij}$ between the classifier in row $i$ and the classifier in column $j$ is defined by
\[
a_{ij} = 100 \cdot\frac{2}{\abs{\S{D}}}\sum_{d \in \S{D}} \cdot \frac{\text{acc}_d(i)-\text{acc}_d(j)}{\text{acc}_d(i) + \text{acc}_d(j)},
\]
Positive (negative) values $a_{ij}$ mean that the average accuracy of the row classifier was higher (lower) on average than the average accuracy of the column classifier.

\end{small}

\bibliographystyle{plain}

\end{document}